%% file: root.tex
\documentclass[letterpaper, 10 pt, conference]{ieeeconf}  

\IEEEoverridecommandlockouts                              

\let\labelindent\relax
\usepackage{amssymb}
\usepackage{graphicx}
\usepackage[hidelinks]{hyperref}
\usepackage[ruled,resetcount,vlined]{algorithm2e}
\usepackage{commath}
\usepackage[font=small]{caption}
\usepackage{subcaption}
\usepackage{booktabs}
\usepackage{diagbox}
\usepackage{amsthm}
\usepackage{times} 
\usepackage{amsmath} 
\usepackage{graphicx}
\usepackage{textcomp}
\usepackage[dvipsnames, table]{xcolor}
\usepackage{float}
\usepackage{wrapfig}
\usepackage{svg}
\usepackage{color, soul}
\usepackage[capitalize]{cleveref}
\usepackage[style=ieee, sorting=none,backend=biber, maxbibnames=2, minbibnames=1]{biblatex}
\usepackage{tikz,subcaption}
\usetikzlibrary{arrows.meta}
\usepackage{enumitem}
\usepackage{lipsum}

\title{Moving On, Even When You're Broken: Fail-Active Trajectory Generation via Diffusion Policies Conditioned on Embodiment and Task \vspace{-10pt}}

\author{Gilberto G.~Briscoe-Martinez$^{*, 1}$, Yaashia Gautam$^{2}$, Rahul Shetty$^{1}$, Anuj Pasricha$^{1}$, \\ Marco M. Nicotra$^{2}$, and Alessandro Roncone$^{1}$\vspace{-10pt}%
\thanks{$^{*}$ Corresponding author. GBM is supported by NASA Space Technology Graduate Research Opportunity Grant 80NSSC22K1211.}%
\thanks{$^{1}$ Dept. of Computer Science, University of Colorado Boulder, CO, USA.}%
\thanks{$^{2}$ Dept. of Electrical, Computer, and Energy Engineering, University of Colorado Boulder, CO, USA.}%
\thanks{Emails: {\tt\small \{firstname.lastname\}@colorado.edu}}%
}

\begin{document}
\maketitle
\thispagestyle{empty}
\pagestyle{empty}

\begin{abstract}
Robot failure is detrimental and disruptive, often requiring human intervention to recover. Our vision is \textsl{fail-active} operation, allowing robots to safely complete their tasks even when damaged. Focusing on \textsl{actuation failures}, we introduce DEFT, a diffusion-based trajectory generator conditioned on the robot’s current embodiment and task constraints. DEFT generalizes across failure types, supports constrained and unconstrained motions, and enables task completion under arbitrary failure. We evaluate DEFT in both simulation and real-world scenarios using a 7-DoF robotic arm. DEFT outperforms its baselines over thousands of failure conditions, achieving a 99.5\% success rate for unconstrained motions versus RRT's 42.4\%, and 46.4\% for constrained motions versus differential IK's 30.9\%. Furthermore, DEFT demonstrates robust zero-shot generalization by maintaining performance on failure conditions unseen during training. Finally, we perform real-world evaluations on two multi-step tasks, drawer manipulation and whiteboard erasing. These experiments demonstrate DEFT succeeding on tasks where classical methods fail. Our results show that DEFT achieves fail-active manipulation across arbitrary failure configurations and real-world deployments.

\end{abstract}



\input{sections/intro}
\input{sections/related_works}

\input{sections/methods}
\input{sections/experiments}

\input{sections/conclusion}%

\AtNextBibliography{\small}
\printbibliography

\clearpage




\end{document}

%% file: sections/intro.tex
\section{Introduction}

The Mars rover Opportunity’s robotic arm first stalled on November 25, 2005; its shoulder joint experienced intermittent motor stalls for years, managed with voltage workarounds and revised stow procedures. Similarly, Curiosity’s drill paused sampling for roughly eighteen months until a hand-designed drilling method restored coring \cite{nasa_science_first_drilled_sample_2018, jpl_opportunity_shoulder_2008}. Cases like these underscore that failures in continuously operating systems are inevitable \cite{factory_efficiency_2020}. Prevailing safety standards mandate \textsl{fail-freeze} behavior, requiring robots to halt when faults are detected \cite{ISO10218-1, ISO13849-1, IEC60204-1}. The result is prolonged downtime and reduced autonomy. 
In this work, we contrast this overly conservative approach with \textsl{fail-active} behavior, a more ambitious (and harder to achieve) reliability goal in which robots maintain functionality, full or reduced, under fault conditions \cite{aero_reliability}. \textbf{Fail-active behavior is how humans operate}: we continue walking on a sprained ankle, writing with a non-dominant hand, going to work when we’ve had our hearts broken. Importantly, achieving fail-active behavior comes with additional complexity. Faults reshape what the robot can physically do: actuation failures (e.g., joint range and velocity restrictions) shrink the feasible workspace and degrade manipulability, electrical faults (e.g., brownouts) reduce control authority, sensor drift can distort state estimates, and end-effector wear (e.g. lowered friction) degrades contact mechanics. Ultimately, these physical changes mean the robot can no longer move the way it originally planned.

\begin{figure}
    \centering
    \includegraphics[width=\columnwidth]{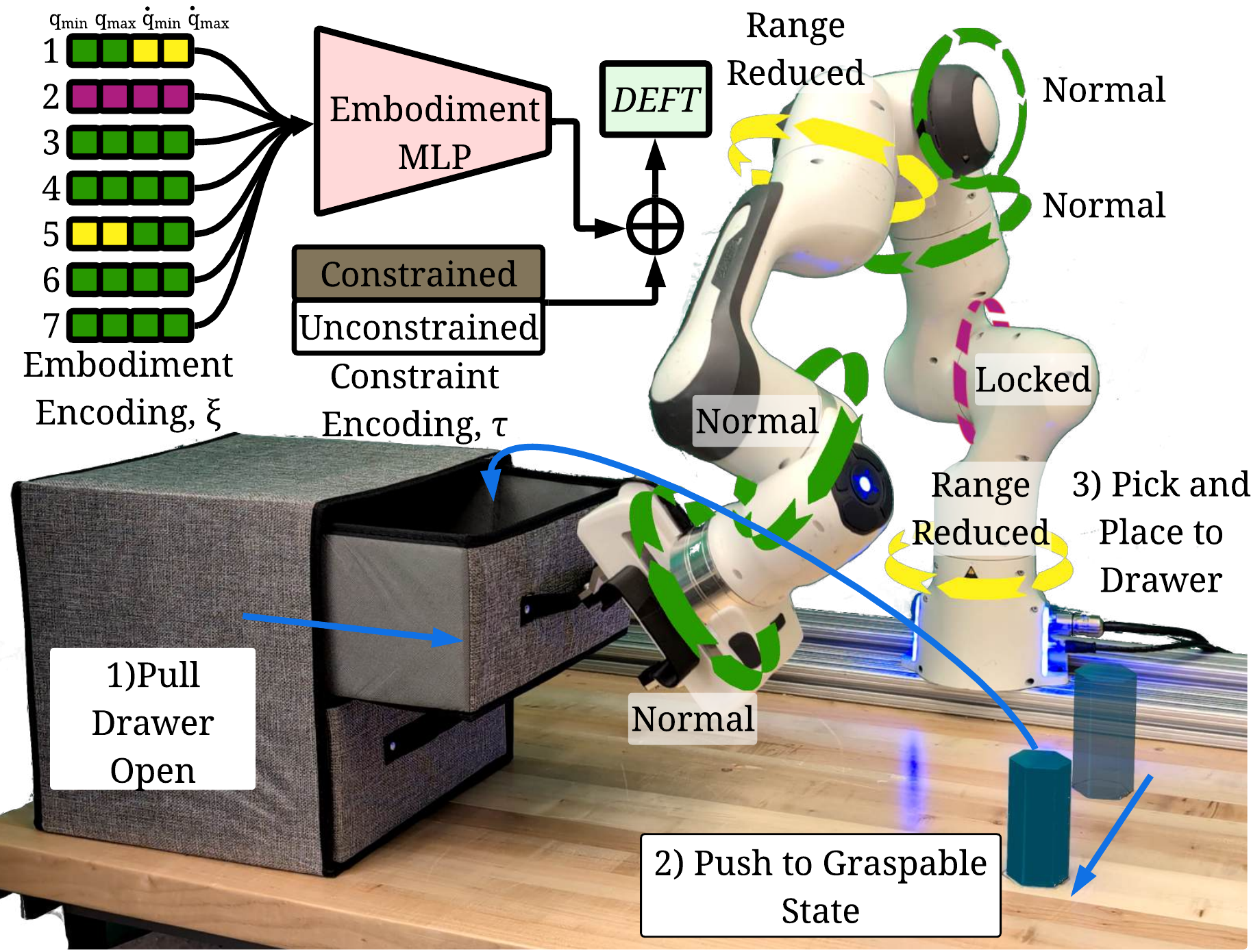 }
    \caption{DEFT takes as input a structured embodiment encoding representing joint-level failures and a constraint encoding then synthesizes a feasible robot trajectory via a diffusion model. After experiencing a failure, a given task will likely need to be completed in a different manner. In this figure, the pick-and-place segment is no longer feasible, given the failure condition of a locked second joint and reduced velocity range on joint 1 and reduced angle range on joint 5, and the robot must now push the object to a graspable state to complete the task. \vspace{-10pt}}
    \label{fig:first}
\end{figure}

In this work, we focus on \textit{actuation failures} as they directly affect how the robot moves. \textit{Fail-active} operation requires adapting to these altered kinematics, which can cause end-effector motion to become non-holonomic; meaning the same control input may no longer yield the same end–effector motion \cite{Bloch_Krishnaprasad_Murray_2016}. 
It is also important to note that since degradation can evolve over time \cite{barbieri2015sensor} and the failure state is not known a priori, \textbf{the space of failure modes is effectively unbounded}. Moreover, each joint can fail independently in multiple ways, and this complexity scales combinatorially with the robot’s degrees of freedom, \textbf{making reliable performance across failures hard to engineer}.

As failure-induced constraints redefine the robot's workspace, manipulability, and feasible interactions with the world, \textsl{we posit that each failure mode can be interpreted as a new embodiment}.  
Through this lens, different embodiments are unable to perform the same tasks using the same behaviors. As no same set of behaviors can span all embodiments, multiple motion primitives are needed to expand the feasible action space and reachable task-space \cite{briscoe2024exploring}.
For example, in \cref{fig:first}, reduced ranges on two joints and one locked joint prevents the robot from grasping the object. However, first pushing the object into a graspable pose and then picking-and-placing it inside the drawer, makes the task feasible.


Synthesizing effective motion primitives for arbitrary failure conditions is challenging due to the complex relationship between robot embodiment constraints and feasible task-space behaviors. Existing approaches fall short in three key gaps: 1) they do not generalize to arbitrary failure configurations \cite{briscoe2024exploring, xie2021maximizing}; 2) they are not capable of completing multiple manipulation primitives \cite{pham2024adaptive}; and 3) they do not address multi-joint failures \cite{mu2016kinematic, kim2024quadrupedal}. To the best of the authors’ knowledge, no existing work bridges all three of these gaps simultaneously. 
Under the interpretation of failure as embodiment shift, these capabilities have not been realized as there are potentially infinite embodiments to generalize across and each embodiment has its own distribution of actions capable of achieving a set task. 
Diffusion models are uniquely suited to solve this. Because they excel at modeling multimodal data distributions, diffusion policies can generate across the range of actions required to span these infinite embodiments at inference.

To this end, we present \textit{DEFT}: a \textit{D}iffusion-based \textit{E}mbodiment-aware \textit{F}ail-active \textit{T}ask-conditioned trajectory generation framework, capable of maintaining robot manipulation functionality under failure conditions. Our contributions are:
(i) an \textit{embodiment vector} encoding per-joint actuation failures for online adaptation to previously unseen failure-induced embodiments and
(ii) a \textit{constraint encoding} selecting between constrained and unconstrained motions to enable the use of multiple motion primitives.
Furthermore, we utilize start–goal inpainting and output clamping to enforce embodiment failure conditions \cite{repaint_lugmayr_2022}.
This allows \textsl{DEFT} to: 1) generalize to arbitrary failure configurations; 2) complete multiple manipulation primitives; and 3) handle multi-joint failures.
We benchmark DEFT against conventional and learning based motion-planning approaches and demonstrate significant improvements in trajectory generation under arbitrary failures (up to 84\% success), adherence to task constraints (up to 78\% success), and generalization to previously unseen failures (up to 99\% success). Collectively, these results enable reliable fail-active operation, allowing robots to maintain functionality and safely complete tasks without human intervention across arbitrary failure conditions.

\vspace{-15pt}

%% file: sections/related_works.tex
\section{Related Work}
\paragraph{Failure-Aware Robot Control}

Robots operating over long durations in uncertain environments must be able to continue functioning even when experiencing physical degradation. Joint-level failures, such as locking, reduced range of motion, or diminished velocity, introduce complex, nonlinear constraints that disrupt standard motion planning approach. Classical methods attempt to address these challenges through explicit algorithmic adaptation: self-motion manifold planning~\cite{xie2021maximizing}, fail-safe reachability analysis~\cite{porges2021failsafe}, and redundancy exploitation via inverse kinematics~\cite{rayankula2021fault}. In assuming a subset of possible failures, these approaches offer guarantees on post-failure ability but they are unable to extend to arbitrary failure conditions. Other works adapt control laws to handle specific joint failures~\cite{yang2024qp, wang2021ik, khan2023chaos}, but are often limited in generality, scale poorly with the combinatorial space of failure types, and rely heavily on manual modeling. While effective in narrow domains, these handcrafted solutions cannot address the full complexity of real-world fail-active behavior, where failures must be handled, and task strategies adapted, on the fly.


Learning-based strategies address the limitations of traditional methods by training control policies that adapt to embodiment changes through experience. In reinforcement learning, damage-aware behaviors have been achieved using adversarial training \cite{yang2021adversarial}, partial observability \cite{pham2024adaptive}, and stochastic joint masking \cite{kim2024quadrupedal}. Other methods apply curriculum learning \cite{okamoto2022acdr} or quality diversity search \cite{cully2015qd, allard2023hierarchical} to develop behavioral repertoires for failure recovery.
These techniques have shown success in recovering locomotion or point-to-point reaching under known embodiments, but often require task-specific training loops, explicit policy switching, or runtime optimization. 
In contrast, our approach trains a generative diffusion model over full trajectories, enabling flexible, zero-shot adaptation to joint degradation and task shifts without policy switching or fine-tuning.

\paragraph{Diffusion Models for Trajectory Generation}
Diffusion models offer key advantages for trajectory generation in fail-active robotic systems, especially when embodiment changes unpredictably. Compared to imitation or reinforcement learning, diffusion models (1) capture complex, multimodal action distributions essential for recovery behaviors \cite{chi2023diffusion}, (2) provide stable training, outperforming alternatives like energy-based models \cite{zhong2023edmp}, and (3) enable conditional generation on structured inputs such as goals or embodiment states \cite{yang2023dimsam}. These features allow robots to adapt to joint failures without explicit retraining.

Empirically, diffusion models excel in robotic tasks like navigation, manipulation, and object rearrangement \cite{sridhar_nomad_2023}, outperforming reinforcement learning and behavior cloning in long-horizon, contact-rich tasks \cite{dasari_ingredients_2024}. Advances in diffusion transformers, like adaptive normalization and efficient tokenization, further enhance their suitability for continuous control \cite{dasari_ingredients_2024}. Crucially, diffusion supports online adaptation via trajectory reconditioning, enabling flexible fail-active responses without policy switching \cite{zhong2023edmp}, making them ideal for handling joint failures that reshape the motion space.

%% file: sections/methods.tex
\section{Methods}
\begin{figure*}
  \centering
  \includegraphics[width=\textwidth]{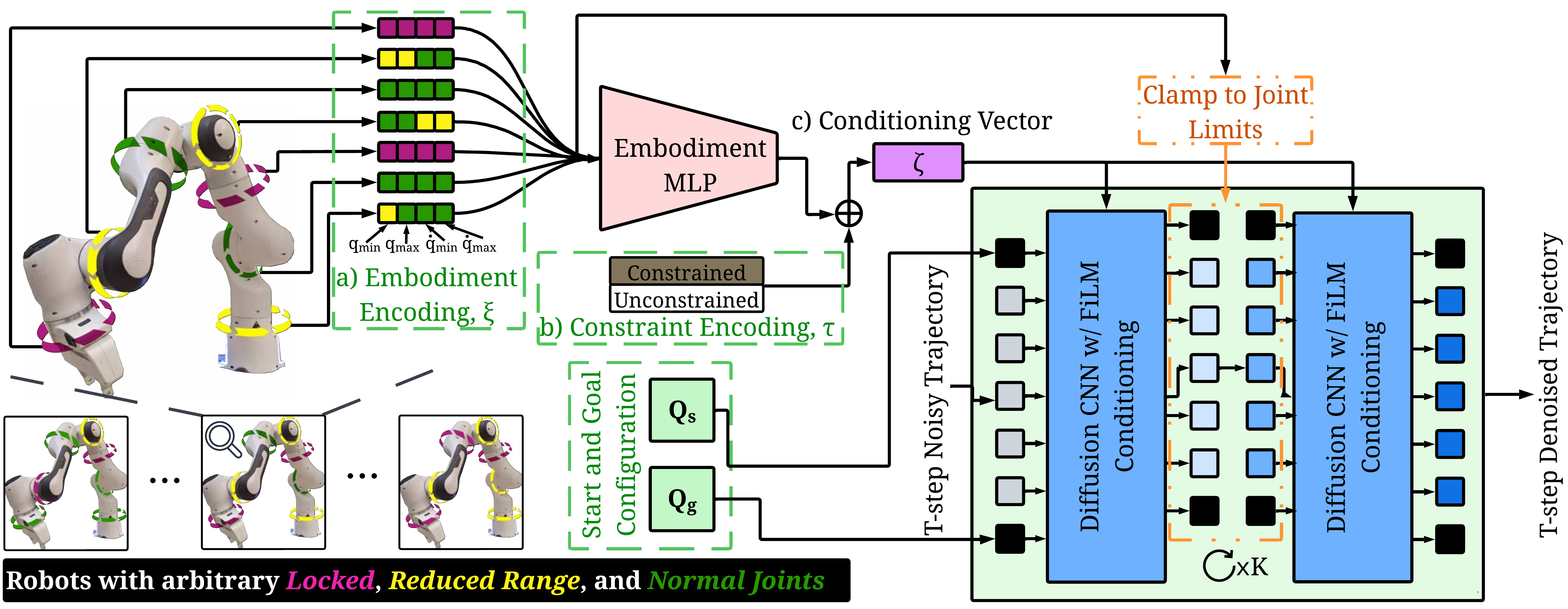}
  \caption{Overview of DEFT. a) joint-level embodiment constraints are encoded into a structured representation capturing failure constrained joint position and velocity limits. This embodiment encoding is processed by a multilayer perceptron (MLP) and used to condition the diffusion model. b) Task-specific constraints (e.g. unconstrained or constrained) are represented as one-hot vectors and concatenated with the embodiment encoding to constrained the generative process. 
  Given these conditioning signals and start-goal joint configurations, the model generates feasible joint-space trajectories adapted explicitly to both the robot's degraded embodiment and task requirements. At each step of the denoising process the predicted joint values are clamped to the failure-induced robot joint limits. \vspace{-12pt}}
  \label{fig:system-diagram}
\end{figure*}




Our goal is fail-active manipulation: generate functional, feasible trajectories under joint-level failures. We pose this as conditional trajectory generation given the failure-induced embodiment and the task constraint. Each failure specifies a new embodiment of the robot, and each constraint (e.g., \texttt{unconstrained} vs. \texttt{constrained} motion) induces a distinct trajectory distribution. Given the start and goal joint configurations  the policy $\pi$ infers a future sequence of joint states over a fixed number of steps $T$. Unlike prior work that assumes a fixed set of failure configurations to control over, our method adapts online. It learns both the control behavior and the allowable motions from the current failure condition and the task constraint.
Our formulation enables zero shot trajectory generation across diverse embodiments and task conditions. For example, when a shoulder joint becomes unusable, the robot can switch from lifting to pushing an object without task specific retraining or policy switching.

\subsection{DEFT}
Inspired by recent advances in diffusion policies~\cite{chi2023diffusion}, we develop DEFT for an $N$-degrees of freedom robot to synthesize joint-space trajectories. Formally, the policy is defined as:
\begin{equation}
\pi(Q_{s,g}| \xi, \tau) \rightarrow \{\textbf{q}_{1:T}, \dot{\textbf{q}}_{1:T}\}
\end{equation}

The policy is conditioned on a vector $\xi \in \mathbb{R}^{4N}$ that encodes joint-level embodiment constraints imposed by the failure condition, and on a task constraint represented by a one-hot encoded vector $\tau$, as shown in \cref{fig:system-diagram}.

Given a ground-truth trajectory $\mathbf{x}$ and a diffusion time step, we first add Gaussian noise $\mathcal{N}(\mathbf{0}, \mathbf{I})$ to obtain a corrupted trajectory. We then inpaint the start and goal joint positions, $Q_{s,g} = [{(\textbf{q}_s, \dot{\textbf{q}}_s), (\textbf{q}_g, \dot{\textbf{q}}_g)}], q \in \mathbb{R}^N$, so endpoints remain exact. We proceed by clamping the entire noisy trajectory to the joint angle and velocity limits specified by the failure embedding $\xi = [\mathbf{e}_q, \mathbf{e}_{\dot{q}}] \in \mathbb{R}^{4N}$, ensuring feasibility under the current embodiment. Next, we form a conditioning vector, $\zeta = [\xi, \tau]^\top $,  by encoding the failure embedding with a small MLP and concatenating the resulting features to the task constraint token. The denoiser (UNet) receives the clamped, inpainted trajectory together with the time step and conditioning vector, and predicts a denoised trajectory $\hat{\mathbf{x}}$, denoising for $K=25$ steps as it showed no change in trajectory quality while having a inference speed improvement compared to \cite{chi2023diffusion}. Finally, we hard-enforce constraint adherence by clamping the prediction to the same limits and re-inpaint the endpoints, returning the denoised, feasible trajectory.

\subsection{Conditioning Approaches}

\input{algorithms/diffusion_policy}
\subsubsection{Embodiment Conditioning}
To inform the model of the robot's failure constrained embodiment, we use an embodiment encoding vector $\xi$ to condition the diffusion model. We model joint-level failures as per-joint constraints on the robot's actuation capabilities. Each failure embodiment is represented as $\xi = [\mathbf{e}_q, \mathbf{e}_{\dot{q}}] \in \mathbb{R}^{4N}$, with $\mathbf{e}_{q,j} = [q_j^{\min}, q_j^{\max}]^\top$ and $\mathbf{e}_{\dot{q},j} = [\dot{q}_j^{\min}, \dot{q}_j^{\max}]^\top$ for all $j \in \{1,\dots,N\}$ such that $\mathbf{0} \in \mathbf{e}_{\dot{q},j}$\footnote{$\mathbf{0}$ must be in $\mathbf{e}_{\dot{q},j}$ to ensure a stable equilibrium point exists~\cite{blanchard_differential_2011}.}.  
 Failures reshape the robot’s feasible configuration space and task-space reachability and exist in a continuous space, allowing $\xi$ to capture a wide spectrum of degradation severities, as shown in \cref{fig:system-diagram}. Let $\{\mathbf{q}_t\}_{t=1}^T$ denote a trajectory over $T$ steps, where $\mathbf{q}_t \in \mathbb{R}^N$ is the joint configuration at time $t$, and $\dot{\mathbf{q}}_t$ its velocity. A configuration $\mathbf{q}_t$ satisfies the feasibility set $\mathcal{C}_t(\mathbf{q}_t,\xi)$ if:%
{\setlength{\abovedisplayskip}{6pt}\setlength{\belowdisplayskip}{6pt}\setlength{\abovedisplayshortskip}{4pt}\setlength{\belowdisplayshortskip}{4pt}%
\begin{equation}
\begin{aligned}
\small
  \mathcal{C}_t(\mathbf{q}_t,\xi)=\Big\{\mathbf{q}_t\in\mathbb{R}^N\ \Big|\ 
  q_{t,j}\in \mathbf{e}_{q,j},\ \dot{q}_{t,j}\in \mathbf{e}_{\dot{q},j}\ \\ \ \forall j\in\{1,\dots,N\}\Big\}.
\end{aligned}
\end{equation}}%
A trajectory is feasible if $\mathbf{q}_{1:T}\in\bigcap_{t=1}^T \mathcal{C}_t(\mathbf{q}_t,\xi)$, i.e. the joint trajectory for the time horizon $T$ should be within joint limits. We condition on failures structurally: the model learns to generate motion plans that obey the constraints.%
To integrate embodiment information into the generative process, we pass $\xi$ through an MLP to produce an embedding that modulates the diffusion model via FiLM. To ensure that the trajectory connects the desired start and goal states we apply start–goal inpainting, fixing $(\mathbf{q}_s,\mathbf{q}_g)$ in the input trajectory. This anchors the trajectory; without it, the stochastic denoiser may miss hard endpoint constraints. In addition, we clamp the trajectory to the limits during both training and inference: conditioning guides the model toward feasibility but does not guarantee it, so clamping enforces satisfaction of the joint constraints.

\subsubsection{Constraint Conditioning}  
We further condition the model on discrete task constraints using a one-hot encoding vector, $\tau$. Let $\tau \in \{0,1\}^K$ be a one-hot vector with exactly one nonzero entry, such that  
\[
\sum_{k=1}^{K} \tau_k = 1 
\quad \text{and} \quad 
\tau_k \in \{0,1\} \ \forall k.
\]  
We concatenate $\tau$ with $\xi$ to form the conditioning vector $\zeta = [\xi, \tau]^\top$, which is injected via FiLM.  

These constraints are not explicitly enforced during inference but must be learned implicitly through the conditioning signal. The constraint encoding also informs the model of task-imposed end effector constraints, which it otherwise would not know while operating in joint space.


\subsection{Data Generation Approaches}
\label{sec:data_generation}
\input{algorithms/push_generation}

We generate a dataset of failure-conditioned joint-space trajectories for two distinct task constraints: \texttt{constrained} and \texttt{unconstrained}. Each trajectory $\mathbf{q}_{1:T} \in \mathbb{R}^{T \times N}$ consists of $T$ joint configurations sampled at fixed intervals of $\Delta t$ seconds. These trajectories are used to train and evaluate our diffusion-based planning framework.

\paragraph{Start and Goal Sampling}
For unconstrained, start and goal configurations $(\mathbf{q}_\text{s}, \mathbf{q}_\text{g})$ are sampled over nominal joint limits and filtered for feasibility using self-collision checks and workspace bounds. For constrained, we instead filter for joint configurations that have planar end-effector poses $(p_\text{s}, p_\text{g})$ at a fixed $z$-height.

\paragraph{Trajectory Construction}
As shown in Algorithm \ref{alg:traj_generation}, Unconstrained trajectories are computed using an RRT-Connect joint-space planner with goal biasing. Constrained trajectories are generated by interpolating end-effector poses along a straight Cartesian path, then solving for each joint configuration using an optimization-based IK solver with pose consistency constraints. All trajectories are smoothed in time using minimum-jerk optimization to produce zero-velocity endpoints and continuous velocity profiles $(\dot{\mathbf{q}}_t)$.

\paragraph{Failure Condition Sampling}
To simulate embodiment degradation, we sample a structured failure vector $\xi$ per trajectory. We first select the number of affected joints, ranging from one to seven, using a modified exponential decay distribution. Assuming  single joint failures are the most common, we fix the probability of a single joint failing at exactly 50\%, while the remaining 50\% probability is distributed among multiple joint failures such that each additional joint failure is half as likely as the one before it. We then sample one of three failure types: joint angle range reduction, velocity limit reduction, or combination. Modified joint limits are computed from the trajectory’s observed values with added uniform noise margins. In particular, for angle range reduction, the joint's feasible position and velocity range are tightened to encompass the observed extrema with an added random buffer  $\epsilon $.



\paragraph{Task Validity Filtering}
For both constraints, only trajectories that obey the constraint-specific requirements are retained. For constrained motions, validity is determined by: 1) total change in end-effector alignment, 2) manipulability above a certain threshold, and 3) path efficiency measured as the length between the two points as compared to the arc length of the generated path. For unconstrained, we verify feasibility using manipulability and goal reachability checks.

%% file: algorithms/diffusion_policy.tex
\begin{algorithm}[t]
\caption{DEFT}
\label{alg:diffusion}
\LinesNumbered 
\KwIn{Denoising Steps $K$; Failure embedding $\mathbf{\xi}$; Primitive encoding $\mathbf{\tau}$}
\KwOut{Denoised trajectory $\hat{\mathbf{x}}$}

Sample noisy trajectory $\tilde{\mathbf{x}} \sim \mathcal{N}(\mathbf{0}, \mathbf{I})$\;

\ForEach{$k \in K$}{
    Inpaint start and goal: $\tilde{\mathbf{x}}_t[\mathrm{start/goal}] \leftarrow Q_{s,g}$\;
    Clamp $\tilde{\mathbf{x}}_t$ to limits specified by $\mathbf{\xi}$\;
    Fuse encodings: $\zeta \gets \mathrm{MLP}_\xi(\mathbf{\xi}) \,\|\, \mathbf{\tau}$\;
    Predict denoised trajectory: $\hat{\mathbf{x}} \gets \mathrm{UNet}(\tilde{\mathbf{x}}_t,\, t,\, \zeta)$\;
    Clamp output $\hat{\mathbf{x}}$ to limits in $\xi$\;
    Inpaint start and goal: $\hat{\mathbf{x}}[\mathrm{start/goal}] \leftarrow Q_{s,g}$\;
}
\Return{$\hat{\mathbf{x}}$}
\end{algorithm}

%% file: algorithms/push_generation.tex


\begin{algorithm}
\small
\caption{Trajectory Generation with Failure Conditioning (Constrained \& Unconstrained)}
\label{alg:traj_generation}
\LinesNumbered
\KwIn{Start-goal pairs $Q_{s,g}$, nominal joint limits $\mathbf{Q}, \dot{\mathbf{Q}}$, primitive encoding $\tau$}
\KwOut{Valid trajectories with associated failure-conditioned limits}

\ForEach{$(\mathbf{q}_{\text{s}}, \mathbf{q}_{\text{g}}) \in Q(s,g)$}{
    \eIf{$\tau = \texttt{constrained}$}{
        $\mathbf{p}_{1:T} \gets \textsc{InterpolateEEPath}(\mathbf{q}_{\text{s}}, \mathbf{q}_{\text{g}})$\;
        $\mathbf{q}_{1:T} \gets \textsc{SolveIKPath}(\mathbf{p}_{1:T})$\;
    }{
        \If{$\tau = \texttt{unconstrained}$}{
            $\mathbf{q}_{1:T} \gets \textsc{RRTConnect}(\mathbf{q}_{\text{s}}, \mathbf{q}_{\text{g}})$\;
        }
    }

    \If{$\textsc{IsValid}(\mathbf{q}_{1:T}, \tau)$}{
        $(\mathbf{q}, \dot{\mathbf{q}}) \gets \textsc{MinJerkOptimize}(\mathbf{q}_{1:T})$\;
        $J_f \gets \textsc{SampleFailureJoints}()$\;
        $f \gets \textsc{SampleFailureType}()$\;
        $\xi\gets \textsc{ApplyFailures}(J_f, f, \mathbf{q}, \dot{\mathbf{q}}, \mathbf{Q}, \dot{\mathbf{Q}})$\;
        $\textsc{StoreTrajectory}(\mathbf{q}, \dot{\mathbf{q}}, \xi, \tau)$\;
    }
}
\end{algorithm}


%% file: sections/experiments.tex
\section{Experimental Evaluation}
We evaluate DEFT in two settings: (i) a \emph{simulation analysis} and (ii) a \emph{real-world comparison}. We ask if a single policy can remain \emph{fail-active} under embodiment degradation. As shown in \cref{sec:sim-results}, \textsc{DEFT} consistently outperforms classical baselines on task success rate while under arbitrary, multi-joint failures and failures not seen during training. The real-world study further examines if these gains translate to end-to-end execution under induced multi-joint failures and primitive switches in real-world scenarios. Together, the results show that DEFT is a single policy that \emph{generalizes to arbitrary failure configurations}, \emph{completes multiple manipulation primitives}, and \emph{handles multi-joint failures}.
 \vspace{-5pt}

\subsection{Simulation Analysis} \label{sec:sim-analysis}
We test three hypotheses to probe DEFT’s generalization to arbitrary joint failures (\textbf{H1} and \textbf{H2}) and to test whether DEFT produces both unconstrained and constrained motions (\textbf{H3}).:

\begin{enumerate}[label=\textbf{H.\arabic*}]
\item Does DEFT obey embodiment constraints of arbitrary actuation failures more often than classical planners?
\item Does DEFT \emph{generalize} to out-of-distribution (OOD) failures?
\item Does DEFT adhere to constraint conditions more than approaches specialized to a single class of constraints?
\end{enumerate}

\subsubsection{Evaluation Preliminaries}\label{sec:eval-prelim}
Below, we define the task constraints geometrically, specify the dataset and baselines, and describe the in-domain (ID)–out-of-domain (OOD) partitioning in joint-limit space. Finally, we define the statistical procedures used to aggregate outcomes.

\paragraph{Evaluation Setup} 
We use the motion planning method from \cref{sec:data_generation} as our baseline, with RRT for \texttt{unconstrained} motions and differential IK with optimization for \texttt{constrained} motions. These baselines do not handle both primitives. In contrast, \textsc{DEFT} generates both within a unified model.

We evaluate on a fixed-base 7-dof Franka Emika Panda arm with the DEFT model inferencing on an NVIDIA RTX 4090. The study covers 4.7k failure conditions; of which 2.9k includes joint angle constraints (reduced and locked) and the rest are joint velocity constraints. For each condition, we randomly select 1–7 joints to be locked, range-limited, or velocity-restricted. Of these conditions, 22\% are in-domain (ID) and 78\% are out-of-domain (OOD). We classify ID/OOD using Mahalanobis (global) and k-NN (local) distances in joint space, labeling cases beyond the 95th percentile of either metric as OOD. Each unique failure condition is tested 100 times, each with 100 start-goal pairs for a total of 4.7M trajectories evaluated. Approximately 5\% of failure conditions render \texttt{constrained} motion infeasible and were excluded.

\begin{figure}
    \vspace{2mm}
    \centering
    \includegraphics[width=0.9\columnwidth]{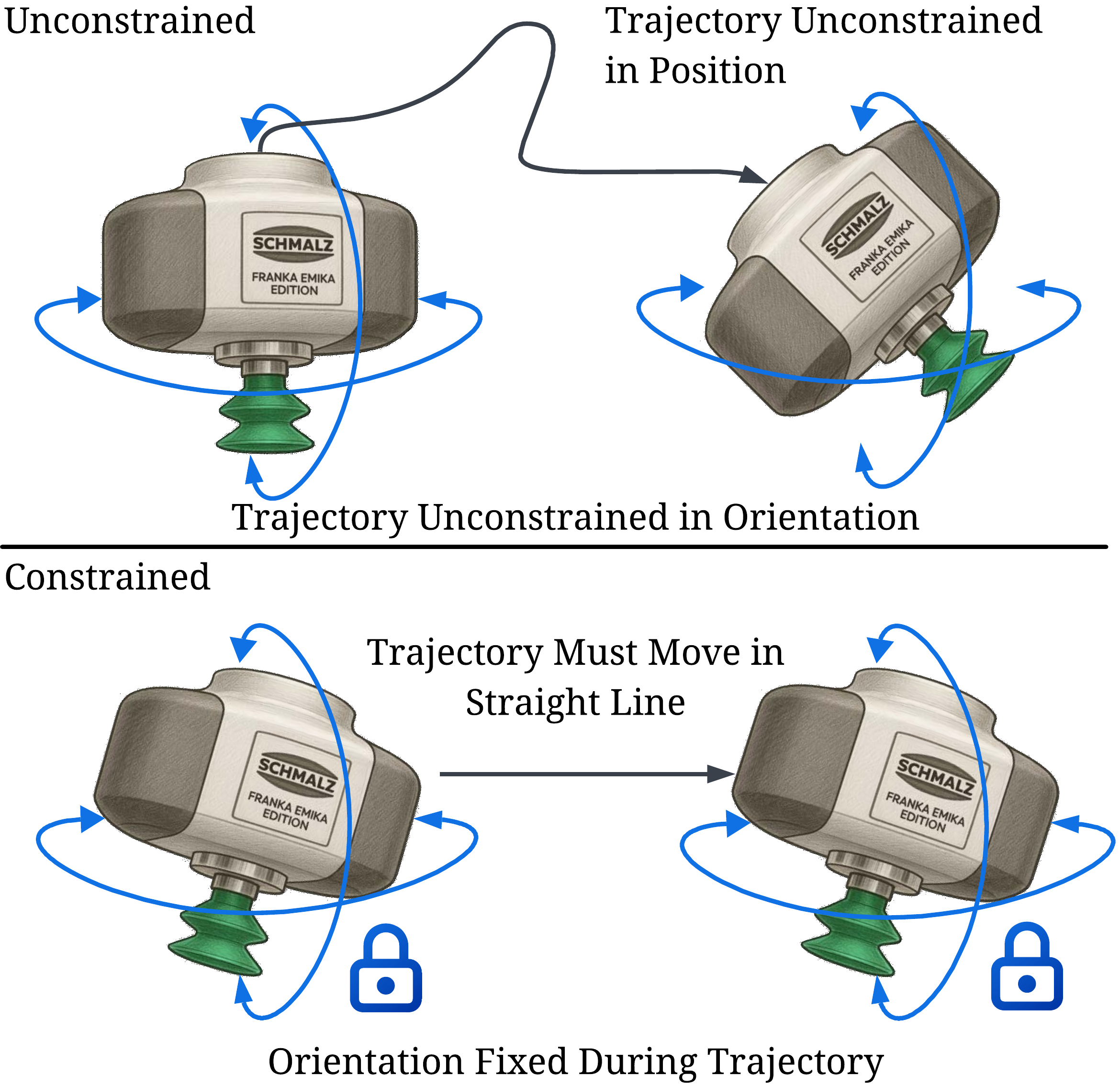 }
    \caption{Graphic description of the constraint definitions used in \cref{sec:sim-analysis}. The \texttt{unconstrained} primitive corresponds to a feasible trajectory between two points and includes unconstrained manipulation where the object is rigidly secured to the end-effector. The \texttt{constrained} corresponds to planar, approximately straight-line motion with minimal end-effector orientation change. \vspace{-25pt}}
    \label{fig:primitive-explain}
\end{figure}

\paragraph{Constraint Definitions}
Constraints correspond to task-relevant motion primitives. Unconstrained motion (\cref{fig:primitive-explain}, top) corresponds to a feasible trajectory between two points and includes manipulation primitives where the object is rigidly secured to the end-effector. The objective is to move from $\mathbf{q}_\text{s}\in\mathbb{R}^N$ to $\mathbf{q}_\text{g}\in\mathbb{R}^N$ with no additional geometric constraints beyond feasibility under failures.%
{\setlength{\abovedisplayskip}{6pt}\setlength{\belowdisplayskip}{6pt}\setlength{\abovedisplayshortskip}{4pt}\setlength{\belowdisplayshortskip}{4pt}%
\begin{equation}\small
\begin{aligned}
  \mathcal{T}_\text{Unconstrained}=\Big\{\mathbf{q}_t \forall t\in\{1,\dots,T\}\ \Big|\ 
  & \mathbf{q}_1=\mathbf{q}_\text{s},\  \mathbf{q}_T=\mathbf{q}_\text{g},\ \\
  & \mathbf{q}_t\in \mathcal{C}_t(\mathbf{q},\xi)\Big\}.
\end{aligned}
\end{equation}}%

Constrained motion (\cref{fig:primitive-explain}, bottom) corresponds to planar, approximately straight-line motion with minimal end-effector orientation change. This provides a spatial constraint for motion primitives such as pushing, pulling, or surface tracing. Let $\mathbf{p}_t\in\mathbb{R}^3$ be the end-effector position, $\mathbf{R}_t\in SO(3)$ its orientation, and $\mathbf{R}_\text{s}$ the initial orientation. Let $\mathcal{P}\subset\mathbb{R}^3$ be the designated plane. A trajectory satisfies \texttt{constrained} if, for all $t$, it obeys: (1) $\mathbf{p}_t\in\mathcal{P}$, (2) $\Delta p_t=\texttt{proj}_{\mathcal{P}}(\mathbf{p}_t)\le \epsilon_p$, (3) $\Delta \mathrm{R}_t=\|\mathbf{R}_t-\mathbf{R}_\text{s}\|_F\le \epsilon_R$, and (4) $\mathbf{q}_t\in \mathcal{C}_t(\xi)$. The valid trajectory set is:%
{\setlength{\abovedisplayskip}{6pt}\setlength{\belowdisplayskip}{6pt}\setlength{\abovedisplayshortskip}{4pt}\setlength{\belowdisplayshortskip}{4pt}%
\begin{equation}\small
\begin{aligned}
  \mathcal{T}_\text{Constrained}
  = \Big\{\mathbf{q}_{t}\ \forall t\in\{1,\dots,T\}\ \Big|\ &
  \mathbf{p}_t\in\mathcal{P}\ \land\ \\ \Delta p_t\le \epsilon_p,
   \Delta \mathrm{R}_t\le \epsilon_R\ & \land\ \mathbf{q}_t\in \mathcal{C}_t(\mathbf{q},\xi)
  \Big\}.
\end{aligned}
\end{equation}}%

\paragraph{Evaluation Metrics}
We applied statistical methods to test whether DEFT outperforms baseline approaches. Because success rates are bounded between 0 and 1 and often vary in spread, we used nonparametric methods that do not assume normally distributed data. To compare DEFT with the baseline under each failure condition, we estimated the difference in success rates using bootstrapping; if the 95\% confidence interval does not include zero, DEFT is considered better. To compare performance across groups of conditions (e.g., in-distribution vs. out-of-distribution, task constraint, or failure type), we used the Mann–Whitney U test, which is robust to unequal variances and outliers. To reduce the risk of false positives when making many comparisons, we applied false discovery rate (FDR) correction. Finally, to assess the overall effect of planner choice, we used a \(\chi^2\)\ test to ask whether using DEFT significantly changes the odds of satisfying constraints across all tasks.

\subsubsection{Results} \label{sec:sim-results}
In this section, we demonstrate that DEFT significantly outperforms traditional planning baselines across various failure conditions. DEFT achieves a 37.66 percentage point improvement in constraint satisfaction compared to classical methods. We also show that DEFT generalizes well to unseen failure conditions, maintaining near-parity performance across in-distribution and out-of-distribution failure conditions. Finally, DEFT excels at handling both constrained and unconstrained motion tasks, with substantial gains in success rates across diverse constraints.

\paragraph{\textbf{H.1} Analysis}%
To evaluate whether DEFT produces trajectories that satisfy failure constraints at a higher rate than traditional planning baselines, we compare overall constraint satisfaction. DEFT achieves a success rate of 74.51\%, while the baseline achieves only 36.85\%. This represents an absolute improvement of 37.66 percentage points. A bootstrap analysis confirms the reliability of this difference ($p < 10^{-10}$), yielding a 95\% confidence interval of [37.21\%, 38.10\%] on the improvement in success rate. The specific performance rates for both angle and velocity failures can be seen in \cref{tab:performance}. Because the improvement holds across heterogeneous failure types and a wide range of sampled magnitudes, it indicates that DEFT generalizes to arbitrary joint-level degradations by conditioning directly on the failure embedding. In practice, given a new failure specification, DEFT translates it into feasible motion plans at substantially higher rates than the baseline.
 \vspace{-5pt}

\begin{table}[h]
\centering
\begin{tabular}{|c|c|c|c|}
\hline
\textbf{Hypothesis Summary} & \textbf{Test Condition}  & \textbf{Baseline} & \textbf{DEFT} \\ \hline
\textbf{H.1} Does DEFT handle & Angle Failure & 48.2\% & \textbf{84.3}\% \\ \cline{2-4}
arbitrary joint failures?& Velocity Failure & 32.5\% & \textbf{70.8}\% \\ \hline
\textbf{H.2} How well does DEFT & In Domain & \cellcolor{black} & \textbf{78.33\%} \\ \cline{2-4}
 handle unseen failures?& Out of Domain & \cellcolor{black} & \textbf{73.61\%} \\ \hline
\textbf{H.3} How well does DEFT  & Constrained & 30.93\% & \textbf{46.42}\% \\ \cline{2-4}
create multiple primitives?  & Unconstrained & 42.4\% & \textbf{99.58\%} \\ \hline

\end{tabular}
\caption{Success rate of the Baseline methods and the DEFT method. \vspace{-10pt}}
\label{tab:performance}
\end{table}


\paragraph{\textbf{H.2} Analysis}
To test whether DEFT generalizes to unseen embodiment failure conditions, we tested success rates across in-distribution (ID) versus out-of-distribution (OOD) embodiment failure conditions, classified using Mahalanobis distance and k-NN distances in joint space. DEFT maintained a high constraint satisfaction rate as shown in \cref{tab:performance}. The small difference between ID and OOD performance indicates robustness to distributional shift. 
Near-parity between ID and OOD success 
indicates DEFT extrapolates over joint space rather than memorizing seen cases. Consequently, previously unseen failure configurations, regardless of which joint is affected or the magnitude of degradation, are translated into geometrically feasible, constraint-satisfying trajectories, evidencing DEFT’s ability to handle arbitrary failure conditions.

\paragraph{\textbf{H.3} Analysis}%
To assess whether DEFT more reliably satisfies task motion constraints, we evaluate performance separately for \texttt{unconstrained} (where the end effector can move freely, such as pick-and-place) and \texttt{constrained} (where the end effector must remain level and move in a straight line, such as surface tracing, pushing, and pulling) motions. The constraint satisfaction rate for both the categories as compared to the baselines (\cref{tab:performance}) shows that DEFT performs better for both primitives.
In the \texttt{unconstrained} setting, DEFT achieves a constraint satisfaction rate of 99.58\%, compared to 42.40\% for the RRT baseline, an absolute improvement of 57.18 percentage points. 
A Mann--Whitney U test on the \texttt{unconstrained} primitive difference between baseline and DEFT confirms that DEFT is statistically significant in 95.24\% of tested conditions after FDR correction, with a median bootstrap 95\% confidence interval on the improvement in success rate of [34.62\%, 79.85\%]. 
For the \texttt{constrained} motions, DEFT achieves 46.42\% satisfaction compared to the baseline's 30.93\%. A chi-squared test confirms that primitive constraint satisfaction is significantly associated with planner choice for both \texttt{unconstrained} and \texttt{constrained} primitives (\(\chi^2\text{ test},\ p<10^{-10}\)) , providing strong empirical support for \textbf{H.3}. It is important to note that the overall low success rate arises from the combined constraints of the primitive and the failure condition. For this evaluation, when excluding conditions, we checked only whether the randomly sampled start and goal satisfied the \texttt{constrained} constraints. However, this check does not guarantee that any feasible action exists, which likely contributes to the sub-50\% success rate.

These findings indicate that \textsc{DEFT} is not specialized to a single manipulation mode: it reliably synthesizes feasible trajectories for both \texttt{unconstrained} and \texttt{constrained} motions under the policy. 

\noindent\emph{Ablation note.} In simulation, conditioning yields similar constraint-obedience to an unconditioned diffusion ablation (within \(\approx\)1–2 percentage points overall), while delivering large gains in real-world task success (\cref{sec:real_world}).
 \vspace{-5pt}

\subsection{Real World}\label{sec:real_world}

Having established in simulation that \textsc{DEFT} generalizes to unseen failures and and reliably instantiates both \texttt{unconstrained} and \texttt{constrained} motions, we now test whether these gains translate to end-to-end task success on hardware. The next subsection evaluates \textsc{DEFT}, the hybrid RRT--Differential-IK optimization approach used in data generation (\cref{sec:data_generation}), and an ablation of DEFT without the conditioning input, \textsc{DEFT}-NoConditioning,  on two long-horizon, multi-primitive tasks (drawer and erasing). \vspace{-15pt}



\begin{table}[h]
\centering
\caption{Applied joint limits (rad) for real-world configurations. \textbf{Bold} indicates a locked joint; \emph{italics} indicate reduced limits.}
\label{tab:limits_drawer_eraser}
\setlength{\tabcolsep}{6pt}\footnotesize
\begin{tabular}{lrr|rr}
\hline
\textbf{Joint} & \multicolumn{2}{c|}{\textbf{Drawer}} & \multicolumn{2}{c}{\textbf{Erasing}} \\
 & \textbf{Lower} & \textbf{Upper} & \textbf{Lower} & \textbf{Upper} \\
\hline
$J_1$ & \emph{-0.81} & \emph{0.17} & \emph{0.033} & \emph{1.028} \\
$J_2$ & \emph{0.10} & \emph{1.50} & -1.763 & 1.763 \\
$J_3$ & \emph{-0.90} & \emph{0.60} & -2.897 & 2.897 \\
$J_4$ & \emph{-2.60} & \emph{-1.30} & \textbf{-2.59} & \textbf{-2.59}\\
$J_5$ & \emph{-0.10} & \emph{1.70} & \emph{-0.073} & \emph{0.957} \\
$J_6$ & \emph{0.90} & \emph{2.50} & \emph{1.865} & \emph{2.958} \\
$J_7$ & \emph{-0.85} & \emph{1.60} & -2.897 & 2.897 \\
\hline
\end{tabular}
\end{table}
\vspace{-10pt}

\begin{figure}[hbtp]
\vspace{2mm}
  \centering
  \includegraphics[width=\columnwidth]{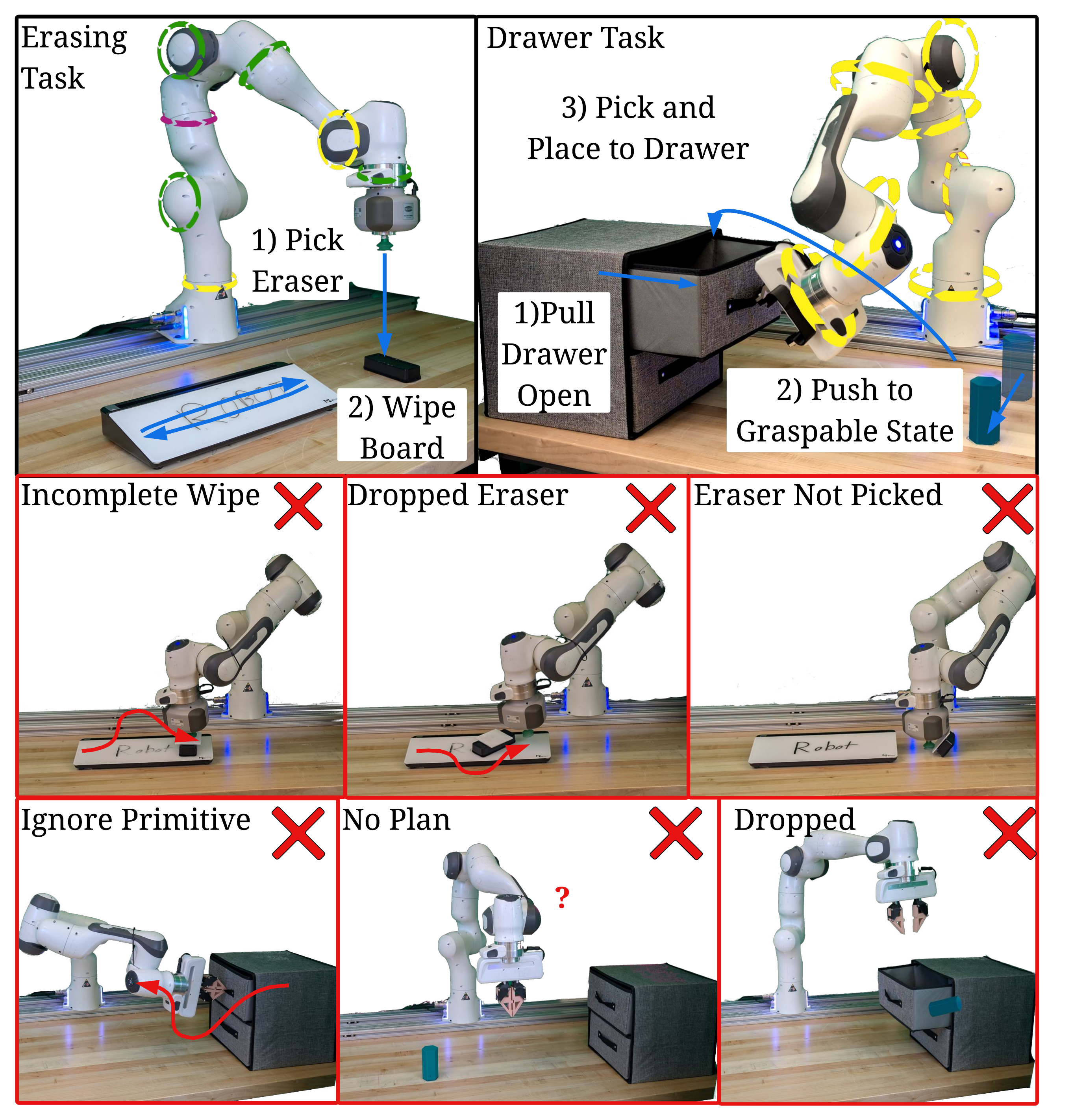}

  \caption{Left (Erasing): The robot grasps an eraser from the table and performs back-and-forth sweeps to remove text. Erasing failures include incomplete wipes or drops from disobeying motion constraints (improper normal force), and missed picks when lacking inpainting prevents reaching prescribed start/goal poses. Right (Drawer): The robot opens a drawer, pushes an object to a graspable pose, places it inside, and closes the drawer. Drawer failures include unopened drawers from ignored motion constraints, unfeasible plans under task constraints, and drops from unrespected target positions without inpainting.\vspace{-23pt}}
  \label{fig:real-world-tasks}
\end{figure}

\subsubsection{Task Description}  
We evaluate our system on two long-horizon, multi-step manipulation tasks that integrate both unconstrained and constrained actions, and that require transitions between unconstrained free-space motion and constrained contact-rich interaction. 

\paragraph{Drawer Task} (\cref{fig:real-world-tasks}, Top Right)
The drawer task consists of five sequential phases: (1) pulling the drawer open from its closed state, (2) pushing the object to a graspable position, (3) grasping an object from the workspace, (4) placing the object inside the drawer, and (5) pushing the drawer closed. The task combines unconstrained manipulation (picking and placing) with constrained manipulation (pushing and pulling), while alternating between unconstrained free-space motion and the constrained kinematics of the drawer. This structure creates distinct primitive requirements and embodiment sensitivities, making it effective for evaluating how conditioning and inpainting influence robustness. 

We assign scores of 1.0 for completing the task and 0.0 for not completing the task as all observed errors led to complete task failure. We evaluated 10 runs per approach.

\paragraph{Erasing Task}  
The erasing task requires the robot to first grasp an eraser from the table, then sweep it back and forth across a whiteboard to remove pre-written text. Here, unconstrained manipulation (picking) leads into constrained, contact-rich surface motion (erasing), requiring sustained alignment with the constrained geometry of the board. This task emphasizes embodiment-dependent failure recovery in a setting where continuous constraint satisfaction and primitive-specific control are necessary, providing a complementary evaluation to the drawer task. 

We partially score each trial up to 1.0 point: (i) +0.25 for picking the eraser, (ii) +0.50 for removing the target text with points awarded only for full removal of markings on the board, and (iii) +0.25 for retaining the eraser through the run. We evaluated 10 runs per approach.

\subsubsection{Failure Condition Description}
For reproducibility, we list the joint limits enforced on the physical robot in \cref{tab:limits_drawer_eraser}. In the drawer task, all seven joints operate under reduced ranges. Collectively, these reductions shrink both reachable and orientation workspaces, diminish null-space freedom, and lower manipulability. In the erasing task, the elbow $J_4$ is locked at $-2.4$\,rad, removing one degree of freedom; $J_1$, $J_5$, and $J_6$ are reduced, further constraining end effector rotation and positioning.

\subsubsection{Results}
In the following section, we present real-world task results that further validate the findings from our simulations. \textsc{DEFT} consistently outperforms traditional baselines, achieving perfect task completion in both the Drawer and Erasing tasks. These results underscore the robustness of \textsc{DEFT} in handling joint-limit failures, showing how embodiment and primitive conditioning enable stable, feasible motion plans where optimization-based methods fail.
\vspace{-15pt}
\begin{table}[h]
\centering
\caption{Real-world task results over 10 runs. }
\label{tab:rw_tasks_combined}
\begingroup
\setlength{\tabcolsep}{5pt}
\renewcommand{\arraystretch}{1.1}
\footnotesize
\begin{tabular}{@{}llcc@{}}
\hline
\textbf{Model} & \textbf{Task} & \textbf{Mean} & \textbf{Std.}  \\
\hline
DEFT                & Drawer  & \textbf{1.00} & 0.00  \\
DEFT                & Erasing & \textbf{1.00} & 0.00 \\
Optimization        & Drawer  & 0.00 & 0.00 \\
Optimization        & Erasing & 0.35 & 0.32 \\
DEFT-NoConditioning & Drawer  & 0.60 & 0.49 \\
DEFT-NoConditioning & Erasing & 0.93 & 0.12 \\
\hline
\end{tabular}
\endgroup
\end{table}
\vspace{-11pt}
\paragraph{Drawer Task}
\textsc{DEFT} achieved perfect task completion on the drawer task (\cref{tab:rw_tasks_combined}). The Optimization baseline produced no executable plans. \textsc{DEFT}-NoConditioning succeeded in \(6/10\) runs. Qualitatively, Optimization repeatedly failed to find feasible trajectories in the constrained drawer setting. The unconditioned ablation failed for two reasons: 1) the lack of embodiment conditioning and 2) missing start--goal inpainting. These missing components caused surface-constraint violations and jerky motion leading to the robot dropping the object and being unable to open the drawer. \textsc{DEFT} alternated cleanly between unconstrained pick-and-place and constrained push and pulls without violating joint or contact limits.

\paragraph{Erasing Task}
\textsc{DEFT} again achieved perfect completion (\cref{tab:rw_tasks_combined}). The Optimization baseline had a success rate of 35\% and exhibited unstable surface interaction with poor contact maintenance. \textsc{DEFT}-NoConditioning had a success rate of 93\% but would disobey the contact constraints causing the robot to lose control of the eraser, consistent with missing embodiment conditioning and inpainting. Under a severe configuration (elbow locked at \(J_4=-2.4\,\mathrm{rad}\) with reduced wrist and base limits), \textsc{DEFT} maintained smooth, constraint-consistent sweeps while preserving grasp.

These outcomes support our central hypothesis: embodiment and task conditioning, coupled with inpainting and clamping, yield a policy that remains fail-active under embodiment degradation. 
Practically, they show that \textsc{DEFT} recovers performance in settings where traditional optimization either cannot plan or produces unstable motions, and where an unconditioned generative policy is brittle. 
Together with the simulation results, the real-world trials substantiate that a single conditioned diffusion policy sustains task success across multiple primitives and joint-limit failures. 


%% file: sections/conclusion.tex
\section{Conclusion \& Future Work} 
One of the practical bottlenecks to long-horizon autonomy is surviving hardware faults without stopping. Building fail-active robots is necessary to maintain functionality under faults without constant human intervention. To that end, we presented \textit{DEFT}, a \textit{D}iffusion-based \textit{E}mbodiment-aware \textit{F}ail-active \textit{T}ask-conditioned trajectory generation framework that enables robots to adapt to joint failures while preserving task performance.
Without explicit replanning or policy switching, DEFT generalizes across multiple failure types and motion primitives. Experiments show DEFT maintains high success rates under multiple joint failures, especially in out-of-distribution scenarios. In real-world trials, DEFT executes reliably under previously unseen multi-joint failures, delivering high task success with strict joint limit compliance.
Future work will focus on real-time failure detection, cross-embodiment transfer to enable fail-active strategies learned on one robot architecture to generalize to different robot platforms, and expanding manipulation skills like pivoting or throwing. By reframing how robots respond to actuation failures, DEFT paves the way to the development of reliable autonomous systems capable of continuous operation under adverse conditions. 
\vspace{-10pt}
